\documentclass[runningheads]{llncs}

\usepackage{graphicx} 
\usepackage{cite}
\usepackage{amsmath,amssymb,amsfonts}
\usepackage{algorithmic}
\usepackage{textcomp}
\usepackage{xcolor}
\usepackage{multirow}
\usepackage{flushend}
\usepackage{booktabs}
\usepackage{comment}
\usepackage[section]{placeins}

\usepackage{amsmath}
\usepackage{subcaption}
\usepackage{graphicx}
\usepackage{caption}
\usepackage{float}
\usepackage{hyperref}
\usepackage{amsmath, tabularx}

\begin{document}

\title{Hierarchical Attentional Hybrid Neural Networks for Document Classification}

\author{Jader Abreu$^{\star}$ \and
 Luis Fred\thanks{Authors contributed equally and are both first authors.} \and
David Mac\^{e}do \and
Cleber Zanchettin
}
%
%
\authorrunning{Abreu, J., Fred, L., et al.}

\institute{\textit{Centro} \textit{de} \textit{Inform\'{a}tica \\
\textit{Universidade} \textit{Federal} \textit{de} \textit{Pernambuco}\\
50.740-560, Recife, PE, Brazil}\\
\email{\{jaoa,lfgs,dlm,cz\}@cin.ufpe.br}}


\maketitle  

\begin{abstract}

Document classification is a challenging task with important applications. The deep learning approaches to the problem have gained much attention recently. Despite the progress, the proposed models do not incorporate the knowledge of the document structure in the architecture efficiently and not take into account the contexting importance of words and sentences. In this paper, we propose a new approach based on a combination of convolutional neural networks, gated recurrent units, and attention mechanisms for document classification tasks. The main contribution of this work is the use of convolution layers to extract more meaningful, generalizable and abstract features by the hierarchical representation. The proposed method in this paper improves the results of the current attention-based approaches.


\end{abstract}


\section{Introduction} 
\vspace{-2mm}
Text classification is one of the most classical and important tasks in the machine learning field. The document classification, which is essential to organize documents for retrieval, analysis, and curation, is traditionally performed by classifiers such as Support Vector Machines or Random Forests. As in different areas, the deep learning methods are presenting a performance quite superior to traditional approaches in this field \cite{c13}. Deep learning is also playing a central role in Natural Language Processing (NLP) through learned word vector representations. It aims to represent words in terms of fixed-length, continuous and dense feature vectors, capturing semantic word relations: similar words are close to each other in the vector space. 

In most NLP tasks for document classification, the proposed models do not incorporate the knowledge of the document structure in the architecture efficiently and not take into account the contexting importance of words and sentences. Much of these approaches do not select qualitative or informative words and sentences since some words are more informative than others in a document. Moreover, these models are frequently based on recurrent neural networks only \cite{c1}. Since CNN has leveraged strong performance on deep learning models by extracting more abundant features and reducing the number of parameters, we guess it not only improves computational performance but also yields better generalization on neural models for document classification.

A recent trend in NLP is to use attentional mechanisms to modeling information dependencies without regard to their distance between words in the input sequences. 
In \cite{c1} is proposed a hierarchical neural architecture for document classification, which employs attentional mechanisms, trying to mirror the hierarchical structure of the document. The intuition underlying the model is that not all parts of a text are equally relevant to represent it. Further, determining the relevant sections involves modeling the interactions and importance among the words and not just their presence in the text.


In this paper, we propose a new approach for document classification based on CNN, GRU \cite{c12} hidden units and attentional mechanisms to improve the model performance by selectively focusing the network on essential parts of the text sentences during the model training. Inspired by \cite{c1}, we have used the hierarchical concept to better representation of document structure. We call our model as Hierarchical Attentional Hybrid Neural Networks (HAHNN). We also used temporal convolutions \cite{c6}, which give us more flexible receptive field sizes. We evaluate the proposed approach comparing its results with state-of-the-art models and the model shows an improved accuracy.\\[-8mm]


\section{Hierarchical Attentional Hybrid Neural Networks}

\begin{figure}[!h]
    \centering
    \includegraphics[width=0.40\textwidth]{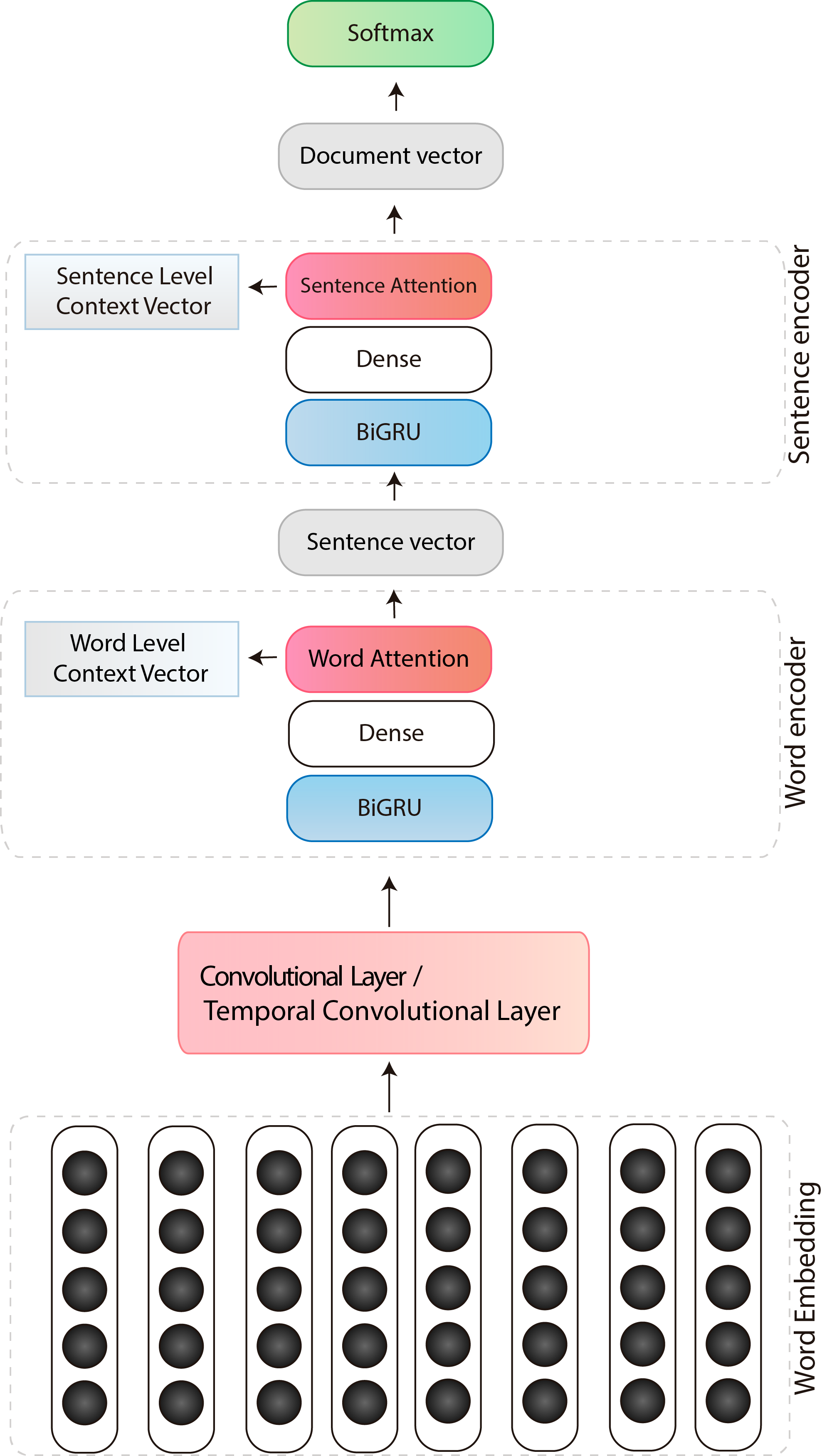}
    \caption{Our HAHNN Architecture include an CNN layer after the embedding  layer. In addition, we have created a variant which includes a temporal convolutional layer \cite{c6} after the embedding layer.}
    \label{fig:fig1}
\end{figure}

The HAHNN model combines convolutional layers, Gated Recurrent Units, and attention mechanisms. Figure~\ref{fig:fig1} shows the proposed architecture. The first layer of HAHNN is a pre-processed word embedding layer (black circles in the Figure~\ref{fig:fig1}). The second layer contains a stack of CNN layers that consist of convolutional layers with multiple filters (varying window sizes) and feature maps. We also have performed some trials with temporal convolutional layers with dilated convolutions and gotten promising results. Besides, we used Dropout for regularization. In the next layers, we use a word encoder applying the attention mechanism on word level context vector. In sequence, a sentence encoder applying the attention on sentence-level context vector. The last layer uses a Softmax function to generate the output probability distribution over the classes. 

We use CNN to extract more meaningful, generalizable and abstract features by the hierarchical representation. Combining convolutional layers in different filter sizes with both word and sentence encoder in a hierarchical architecture let our model extract more rich features and improves generalization performance in document classification. To obtain representations of more rare words, by taking into account subwords information, we used FastText \cite{c3} in the word embedding initialization.

We investigate two variants of the proposed architecture. There is a basic version, as described in Figure~\ref{fig:fig1}, and there is another which implements a TCN \cite{c6} layer. The goal is to simulate RNNs with very long memory size by adopting a combination of dilated and regular convolutions with residual connections. Dilated convolutions are considered beneficial in longer sequences as they enable an exponentially larger receptive field in convolutional layers. More formally, for a 1-D sequence input $\mathrm{ x \in \mathbb{R}^{\it{n}} }$ and a filter $f : \{0,..., k-1\}     \rightarrow \mathbb{R}$, the dilated convolution operation \textit{F} on element \textit{s} of the sequence is defined as \\[-8mm]

\begin{equation}
    F(s) = (x*_df)(s) = \sum_{i=o}^{k-1}f(i) \cdot \bf{x}_{\it{s - d \cdot i}}
\end{equation}

\vspace{-1mm}

where \textit{d} is the dilatation factor, \textit{k} is the filter size, and $\it{s - d \cdot i}$ accounts for the past information direction. Dilation is thus equivalent to introducing a fixed step between every two adjacent filter maps. When \textit{d} = 1, a dilated convolution reduces to a regular convolution. The use of larger dilation enables an output at the top level to represent a wider range of inputs, expanding the receptive field. \\[-9mm]

\vspace{-6mm}

The proposed model takes into account that the different parts of a document have no similar relevant information. Moreover, determining the relevant sections involves modeling the interactions among the words, not just their isolated presence in the text. Therefore, to consider this aspect, the model includes two levels of attention mechanisms \cite{c4}. One structure at the word level and other at the sentence level, which let the model pay more or less attention to individual words and sentences when constructing the document representation.

The strategy consists of different parts: 1) A word sequence encoder and a word-level attention layer; and 2) A sentence encoder and a sentence-level attention layer. In the word encoder, the model uses bidirectional GRU \cite{c4} to produce annotations of words by summarizing information from both directions. Therefore, it incorporates the contextual information in the annotation. The attention levels let the model pay more or less attention to individual words and sentences when constructing the representation of the document \cite{c1}.

Given a sentence with words $\it{w}_{it},t \in [0, T]$ and an embedding matrix $\it{W_e}$, a bidirectional GRU contains the forward $GRU \overrightarrow{f}$ which reads the sentence $s_i$ from $w_{i1}$ to $w_{iT}$ and a backward $GRU \overleftarrow{f}$ which reads from  $w_{iT}$ to $w_{i1}$:




\noindent
\begin{tabularx}{\linewidth}{XX}
    \begin{equation}
        \it{x_{it}} = W_ew_{it}, t \in [1,T], 
    \end{equation}
    &
    \begin{equation}
      \overrightarrow{h_{it}} = \overrightarrow{GRU}(\it{x_{it}}), t \in [1,T],
    \end{equation}
    
\end{tabularx}

\vspace{-5mm}

\begin{equation}
    \overleftarrow{h_{it}} = \overleftarrow{GRU}(\it{x_{it}}), t \in [T,1].
\end{equation}

An annotation for a given word $\it{w_{it}}$ is obtained by concatenating the forward hidden state and backward hidden state, i.e., $\it{h_{it}} = [\overrightarrow{h_{it}}, \overleftarrow{h_{it}}]$, which summarizes the information of the whole sentence. We use the attention mechanism to evaluates words that are important to the meaning of the sentence and to aggregate the representation of those informative words into a sentence vector. Specifically,\\[-6mm]

 
\noindent
\begin{tabularx}{\linewidth}{@{}XX@{}}
    \begin{equation}
      \it{u_{it}} = \tanh(\it{W_wh_{it}} + \it{b_w})
    \end{equation}
    &
    \begin{equation}
      \alpha_{it} = \dfrac{\exp(u_{it}^ \top u_w)}{\sum_t \exp(u_{it}^ \top u_w)}
    \end{equation}
\end{tabularx}

\vspace{-5mm}

\begin{equation}
    s_i = \sum \alpha_{it} h_{it}
\end{equation}

\vspace{-2mm}

The model measures the importance of a word as the similarity of $\it{u_{it}}$ with a word level context vector $\it{u_w}$ and learns a normalized importance weight $\alpha_{it}$ through a softmax function. After that, the architecture computes the sentence vector $s_i$  as a weighted sum of the word annotations based on the weights. The word context vector $u_w$ is randomly initialized and jointly learned during the training process.

Given the sentence vectors $s_i$, and the document vector, the sentence attention is obtained as:\\[-6mm]


\noindent
\begin{tabularx}{\linewidth}{XX}
    \begin{equation}
      \overrightarrow{h_{it}} = \overrightarrow{GRU}(\it{s_{i}}), i \in [1,L],
    \end{equation}
    &
    \begin{equation}
      \overleftarrow{h_{it}} = \overleftarrow{GRU}(\it{s_{i}}), i \in [L,1].
    \end{equation}
    
\end{tabularx}

\vspace{-4mm} 

The proposed solution concatenates $h_i = [\overrightarrow{h_i}, \overleftarrow{h_i}]$ $h_i$ which summarizes the neighbor sentences around sentence $\it{i}$ but still focus on sentence $\it{i}$. To reward sentences that are relevant to correctly classify a document, the solution again use attention mechanism and introduce a sentence level context vector $u_s$ using it to measure the importance of the sentences: \\[-7mm]


\noindent
\begin{tabularx}{\linewidth}{XX}
    \begin{equation}
      \it{u_{it}} = \tanh(\it{W_sh_{i}} + \it{b_s})
    \end{equation}
    &
    \begin{equation}
     \alpha_{it} = \dfrac{\exp(u_{i}^ \top u_s)}{\sum_i \exp(u_{i}^ \top u_s)} 
    \end{equation}
    
\end{tabularx}

\vspace{-5mm}

 \begin{equation}
     v = \sum \alpha_{i} h_{i}
 \end{equation}

\vspace{-1mm}

In the above equation, \textit{v} is the document vector that summarizes all the information of sentences in a document. Similarly, the sentence level context vector $u_s$ can be randomly initialized and jointly learned during the training process. The output of the sentence attention layer feeds a fully connected softmax layer. It gives us a probability distribution over the classes. The proposed method is openly available in the github repository \footnote{https://github.com/luisfredgs/cnn-hierarchical-network-for-document-classification}.\\[-8mm]

\section{Experiments and Results}
We evaluate the proposed model on two document classification datasets using 90\% of the data for training and the remaining 10\% for tests. We split documents into sentences and tokenize each sentence. The word embeddings have dimension 200 and we use Adam optimizer with a learning rate of 0.001. The datasets used are the IMDb Movie Reviews \footnote{http://ai.stanford.edu/~amaas/data/sentiment/} and Yelp 2018 \footnote{https://www.yelp.com/dataset/challenge}. The former contains a set of 25k highly polar movie reviews for training and 25k for testing, whereas the classification involves detecting positive/negative reviews. The latter include users ratings and write reviews about stores and services on Yelp, being a dataset for multiclass classification (ratings from 0-5 stars). Yelp 2018 contains around 5M full review text data, but we fix in 500k the number of used samples for computational purposes.\\[-12mm]

\begin{table}[!h]
\caption{Results in classification accuracies.}
\centering
\begin{tabular}{@{}p{0.28\textwidth}*{2}{p{\dimexpr0.37\textwidth-2\tabcolsep\relax}}@{}}
\toprule
\multicolumn{1}{c}{Method} & \multicolumn{2}{c}{Accuracy on test set}  \\
\cmidrule(r{4pt}){2-3} 
& Yelp 2018 (five classes) & IMDb (two classes)  \\
\midrule
VDNN \cite{c31} & 62.14 & 79.47 \\
HN-ATT  \cite{c1} & 72.73 &  89.02 \\
CNN  \cite{c13} & 71.81 &  91.34 \\
Our model with CNN & \textbf{73.28} & 92.26\\
Our model with TCN & 72.63 & \textbf{95.17} \\
\bottomrule
\end{tabular}
\label{tabela1}
\end{table}

\vspace{-4mm}

Table~\ref{tabela1} shows the experiment results comparing our results with related works. Note that HN-ATT \cite{c1} obtained an accuracy of 72,73\% in the Yelp test set, whereas the proposed model obtained an accuracy of 73,28\%. Our results also outperformed CNN \cite{c1} and VDNN \cite{c31}. We can see an improvement of the results in Yelp with our approach using CNN and varying window sizes in filters. The model also performs better in the results with IMDb using both CNN and TCN.

\subsection{Attention Weights Visualizations}

\begin{figure*}[!ht]
  \center
  \begin{subfigure}{\linewidth}
    \center
      \includegraphics[scale=0.60]{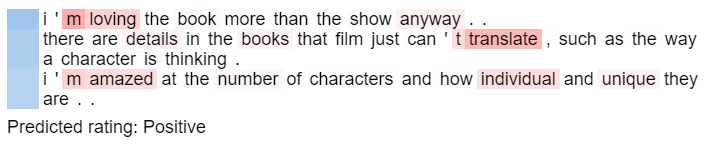}
      \caption{A positive example of visualization of a strong word in the sentence.}  
  \end{subfigure}\par\medskip
  
  \begin{subfigure}{\linewidth}
   \center
   \includegraphics[scale=0.60]{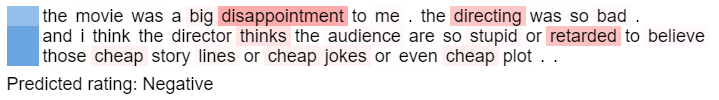}
   \caption{A negative example of visualization of a strong word in the sentence.}
  \end{subfigure}\par\medskip
\vspace{-2mm}
  \caption{Visualization of attention weights computed by the proposed model}
  \label{fig:fig2}
\end{figure*}

To validate the model performance in select informative words and sentences, we present the visualizations of attention weights in Figure \ref{fig:fig2}. There is an example of the attention visualizations for a positive and negative class in test reviews. Every line is a sentence. Blue color denotes the sentence weight, and red denotes the word weight in determining the sentence meaning. There is a greater focus on more important features despite some exceptions. For example, the word ``loving" and ``amazed" in Figure \ref{fig:fig2} (a) and ``disappointment" in Figure \ref{fig:fig2} (b).

\vspace{-4mm}

Occasionally, we have found issues in some sentences, where fewer important words are getting higher importance. For example, in Figure \ref{fig:fig2} (a) notes that the word ``translate" has received high importance even though it represents a neutral word. These drawbacks will be taken into account in future works.\\[-7mm]

\section{Final Remarks}

In this paper, we have presented the HAHNN architecture for document classification. The method combines CNN with attention mechanisms in both word and sentence level. HAHNN improves accuracy in document classification by incorporate the document structure in the model and employing CNN's for the extraction of more abundant features. \\[-8mm] 



\end{document}